\documentclass[10pt,twocolumn,letterpaper]{article}

\usepackage{cvpr}
\usepackage{times}
\usepackage{epsfig}
\usepackage{graphicx}
\usepackage{amsmath}
\usepackage{amssymb}
\usepackage{color}
\usepackage{soul}
\usepackage{flushend}

\cvprfinalcopy 

\def\cvprPaperID{7} 

\ifcvprfinal\fi
\begin{document}

\title{Real-Time Anomaly Detection and Localization in Crowded Scenes}


\author{Mohammad Sabokrou$^1$, Mahmood Fathy$^2$, Mojtaba Hoseini$^1$, Reinhard Klette$^3$\\
$^1$Malek Ashtar University of Technology, Tehran, Iran\\
$^2$Iran University of Science and Technology, Tehran, Iran\\
$^3$Auckland University of Technology, Auckland, New Zealand\\
}

\maketitle

\begin{abstract}
In this paper, we propose a method for real-time anomaly detection and localization in crowded scenes. Each video is defined as a set of non-overlapping cubic patches, and is described using two local and global descriptors. These descriptors capture the video properties from different aspects. By incorporating simple and cost-effective Gaussian classifiers, we can distinguish normal activities and anomalies in videos. The local and global features are based on structure similarity between adjacent patches and the features learned in an unsupervised way, using a sparse auto-encoder.  Experimental results show that our algorithm is comparable to a state-of-the-art procedure on UCSD ped2 and UMN benchmarks, but even more time-efficient. The experiments confirm that our system can reliably detect and localize anomalies as soon as they happen in a video. 
\end{abstract}

\section{Introduction}

The definition of an {\it anomaly} depends on what context is of interest.  A video event is considered as being an anomaly 
if it is not very likely to occur in the video~\cite{CON2011}. 
Describing unusual events in complex scenes is a cumbersome task, often 
solved by employing high-dimensional features and descriptors.  Developing a reliable model to be trained with such descriptors is quite challenging and requires an enormous amount of training samples; it is also of large computational complexity. Therefore, this might face the so-called  ``curse of dimensionality'', in which the predictive power of the trained model reduces, as the dimensionality of the feature descriptors increases.

In recent work, one or a set of reference normal models are learned from training videos, which are then applied for detecting an anomaly in the test phase. Such methods usually consider a test video as being an anomaly if it does not resemble the learned model(s). In order to build these reference models, some specific feature descriptors should be used. In general, features usually are extracted to represent either (1) trajectories or (2) spatio-temporal changes. For instance, \cite{JIA2011} and \cite{WU2010} focus on the trajectories of objects in videos, in which each object is to be labeled as an anomaly or not, based on how they follow the learned normal trajectory. These methods could not handle the occlusion problem, and are also computationally very expensive, for the case of crowded scenes. 

To overcome these weaknesses, researchers proposed methods using low-level features such as optical flow or gradients. They learn the shape and spatio-temporal relations using low-level features distributions. As an example, \cite{MAH2010} fits a Gaussian mixture model as the features, while \cite{ADA2008} uses an exponential distribution. 

Clustering of test data using low-level features is exploited in \cite{SAL2012}. In \cite{BEN2009,KIM2009,KRA2009,ZHA2005}, the normal patterns were fitted to a Markov random field,  and  \cite{MEH2009, WAN2007} apply latent Dirichlet allocations. \cite{LI2014} introduces a joint detector of temporal and spatial anomalies, where the authors use a mixture of dynamic textures (MDT) model.

In recent studies, sparse representations of events~\cite{CON2011,CON2013,LU2013} in videos is being heavily explored. Notably, the proposed models in  \cite{CON2011,CON2013,LI2014,MEH2009,ROS2013,LU2013} achieve favorable performance in anomaly detection, however they normally fail in the task of anomaly localization. All these methods, except \cite{LU2013}, are not designed for real-time applications and commonly fail in real-world anomaly detection problems.

In this paper, we propose to represent videos from two different aspects or views, and thus two partially independent feature descriptors. Then, we introduce an approach for integrating these views in a testing step to simultaneously perform anomaly {\it detection} and {\it localization}, in real-time. Unlike previous work, instead of using low-level features, we propose to learn a set of representative features, based on auto-encoders \cite{VIN2008}. 

Our detection framework identifies an anomaly in a real-time manner. Our anomaly detection method has high true-positive and low false-positive rates which make it quite reliable. We evaluate our anomaly detection and localization framework on popular datasets and report the running time for the whole procedure. The comparison with state-of-the-art methods shows the superiority of our method, both in terms of performance and running time. 

The main contributions of our work are as follows: (1) Presenting a feature learning procedure for describing videos for the task of video anomaly localization. This method is time-consuming for training, but the learned features are very discriminative to model the normal patches. (2) Introducing a descriptor-based similarity metric between adjacent patches for detecting sudden changes in spatio-temporal domains. (3) Representing video patches from two different aspects or views. Both local and global feature sets are used for each view. In the final decision, these views support each other. (4) Modeling all normal patches with Gaussian distributions. For a test video, the Mahalanobis distance is used to figure out its relevance for the normal patches. (5) Being real-time, we are able to detect and localize anomalies soon after they occur in a test video or stream.

\begin{figure}[h!]
\begin{center}
  \includegraphics[width=1\linewidth]{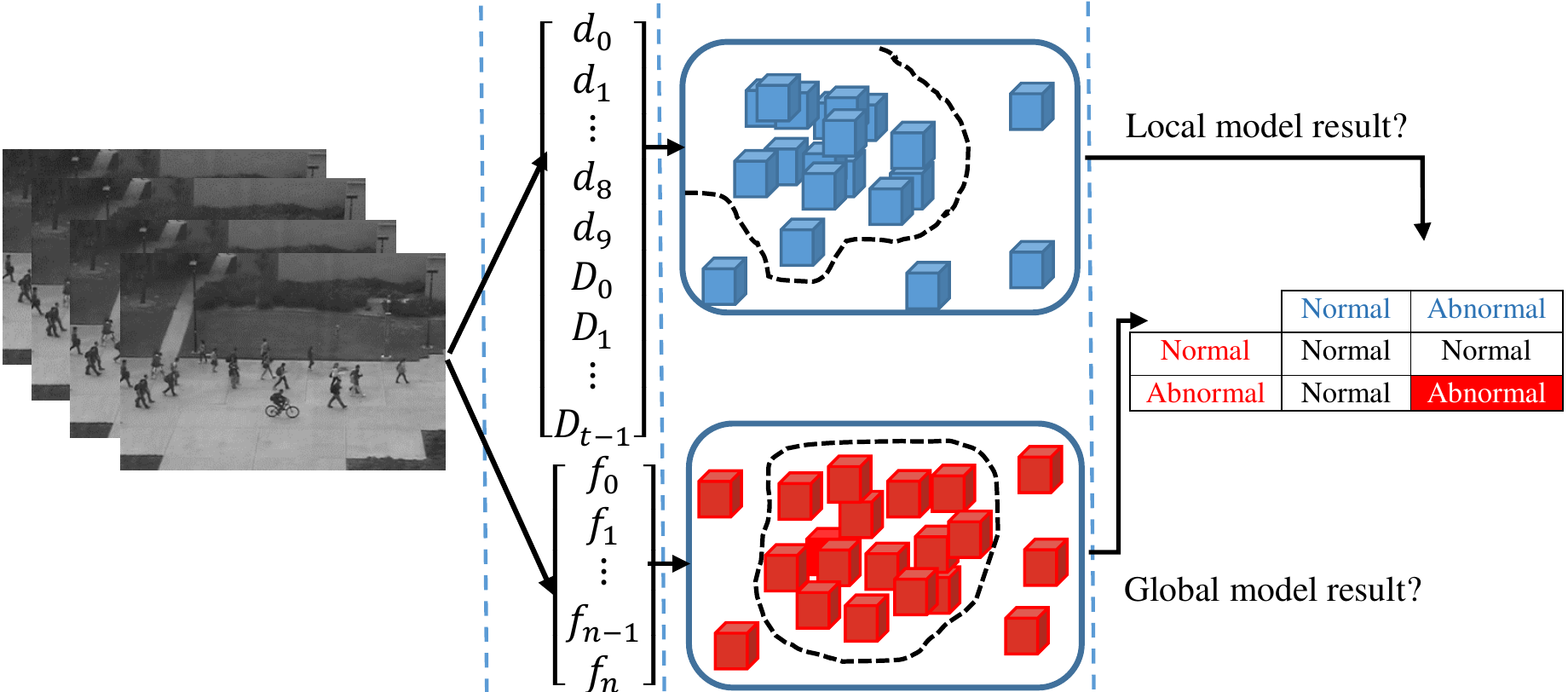}
\end{center}
   \caption
   {
The scheme of our algorithm ({\it left to right}): Input frames,
   two views of patches (global and local), modeling the data using 
   Gaussian distributions, and making the final decision
   }
\label{fig:2}
\end{figure}

The overall scheme of our algorithm is shown in  Fig.~\ref{fig:2}. We achieve 25 fps processing power, and with enduring some bit errors we reach up to 200 fps using a PC with 3.5 GHz CPU and 8G RAM in MATLAB 2012a.

The rest of the paper is organized as follows. The proposed approach is introduced in Section~\ref{sec:PS}, where we first introduce the overall schema, and then we focus on global descriptors, local descriptors, anomaly classification scheme, and finally anomaly detection through feature learning, one after the other. Experimental results, comparisons, and analysis are presented in Section~\ref{sec:ER}. Ultimately, Section~\ref{sec:C}  concludes the paper.

\section{Proposed System}
\label{sec:PS}

{\bf Overall Scheme.}
To represent each video, first each video is converted into a number of non-overlapping cubic patches; a sketch of this video representation is shown in Fig.~\ref{fig:VP}. Generally, every video has one or a set of dominant events. Thus, one expects that {\it normal patches} have similar relations with their adjacent patches and a high likelihood of occurrence in the video. Therefore, these {\it anomaly patches} should meet three conditions:

\begin{figure}[t!]
\begin{center}
  \includegraphics[width=0.8\linewidth]{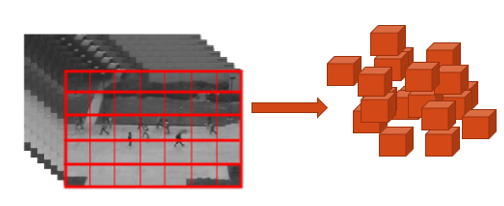}
\end{center}
   \caption{Video representation: Each video is represented through a number of non-overlapping cubic patches, covering the whole space-time in the video.}
\label{fig:VP}
\end{figure}

\vspace{-5pt}
\begin{enumerate}
\itemsep0em 
\item The similarity between the anomaly patches and their adjacent (\ie, defined by spatial changes) 
patches does not follow the same pattern as from normal patches to their adjacent patches.
\item It is most likely that the temporal changes of an anomaly patch would not follow the pattern in the temporal changes of normal patches.
\item It is obvious that the occurrence likelihood of an anomaly patch is less than that of normal patches.
\end{enumerate}
\vspace{-5pt}

It can be easily inferred that the above conditions 1 and 2 are characterized locally. Therefore, they can be encoded by local feature descriptors, and condition 3 is analogous to the global nature of the scene. In other words, conditions 1 and 2 consider the relation between a patch and its adjacent patches, and condition 3 describes the overall appearance of patches in the video. As a result, the first two conditions are corresponding to the spatio-temporal changes, while the latter one is different. Therefore, we model a combination of 1 and 2 through a {\it local} representation, and 3 by a more {\it global} one. On the other hand, in order to avoid the so-called ``curse of dimensionality'', we model these two aspects independently. 

\begin{figure*}[t!]
\begin{center}
  \includegraphics[width=0.8\linewidth]{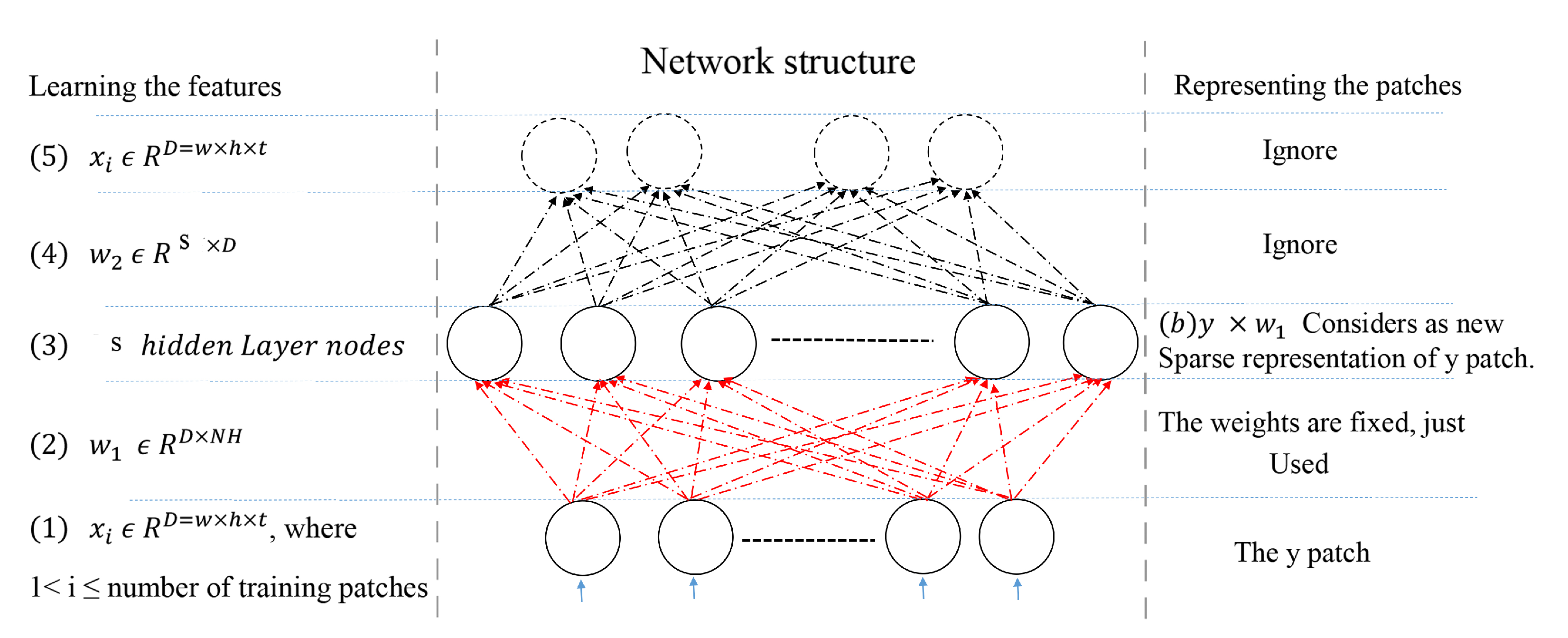}
\end{center}
   \caption
   {
   Summary for learning the global features using an auto-encoder. {\it Left}: The step for
   learning features uses raw normal patches; components (1), (2), (3), (4), and (5)   
   are needed; the aim is to reconstruct the input paths with adjusting $W_{1}$  and  
   $W_{2}$ using gradient descent. {\it Middle}: Auto-encoder structure.
   {\it Right}: Representing the $y$ patch using the $W_{1}$ weights 
   ($y \times W_{1}$); (1), (2), and (3) are just used; this is a multiplication of 
   two matrixes, so it is very fast
   }
\label{fig:AUTO}
\end{figure*}

So far, we have defined two different aspects that we approach the problem, leading to two independent models. In order to make a final decision, we aggregate the decisions from both models. If both models reject a patch it is considered to be an anomaly. This leads to a system with better performance in terms of true-positive and false-positive, since this way of combination of the two models guarantees a concrete selection of a patch as anomaly if both models agree on its being an anomaly. 

In summary, the input videos are represented in two different aspects. Then, these representations are fitted to a set of Gaussian distributions and a decision boundary is calculated for each of them. Finally, based on global and local model results, a decision is reached about a patch being an anomaly or not ({\it detection}). The {\it localization} could be then easily inferred, based on which patches throughout the video are classified as anomaly. In the subsequent sections, the two sets of features (global and local) are introduced.

\vspace{1mm}
{\bf Global descriptors.}
A video global descriptor is a set of features that describes the video as a whole and therefore is best able to describe the normal video patches. In \cite{YAN2013} it is argued that classical handcrafted low-level features, such as HOG and HOF, may not be universally suitable and discriminative enough for every type of video. So, unlike previous works, that use low-level features, we use an unsupervised feature learning method based on auto-encoders. The structure of the auto-encoder is depicted in Fig.~\ref{fig:AUTO}.

The auto-encoder learns sparse features based on gradient descent, by modeling a neural network. Suppose that we have $m$ normal patches with the dimensions $(w, h, t)$, creating a data structure of $x_{i}\in \mathbb{R}^{D}, D=w\times h \times t$ (the raw data). The auto-encoder minimizes the objective defined in Eq.~\eqref{eq:L} by re-reconstructing the original raw data:
\begin{equation} \begin{aligned}
L = &\frac{1}{m}\sum_{i=1}^{m}\Vert x_{i}-W_{2}\delta(W_{1}x_{i}+b_{1})+b_{2}\Vert^{2}\\
&+ \sum_{i=1}^{w\cdot h\cdot  t}\sum_{j=1}^{s}(W_{ji}^2) + \beta \sum_{j=1}^{s}  KL(\rho \Vert \rho^{\prime}_{j})
\label{eq:L}
\end{aligned} \end{equation} 
where $s$ is the number nodes in the auto-encoder's hidden layer, $W_{1}\in\Bbb{R}^{s \times D}$ and $W_{2}\in\Bbb{R}^{D \times s} $ are the weight matrices, which map the input layer nodes to hidden layer nodes, and hidden layer nodes to the output layer nodes, respectively. $W_{ji}$  is the weight between the $j^{th}$ hidden layer node and the $i^{th}$ output layer node, and $\delta$ is equal to the sigmoid function. Furthermore,  $b_{1}$ and $b_{2}$, are the  
bias of the output layer and the hidden layer, respectively. $KL(\rho \Vert \rho^{\prime}_{j})$ is  a regularization function and is set to enforce the activation of the hidden layer to be sparse. $KL$ is based on the similarity between a Bernoulli distribution with $\rho$ as parameter, and  the active node distribution. The parameter $\beta$ is the weight of the penalty term (in the sparse auto-encoder objective). We can efficiently optimize the above objective with respect to $W_{1}$ via the stochastic gradient descent approach.

\begin{figure}[b!]
\begin{center}
  \includegraphics[width=0.8\linewidth]{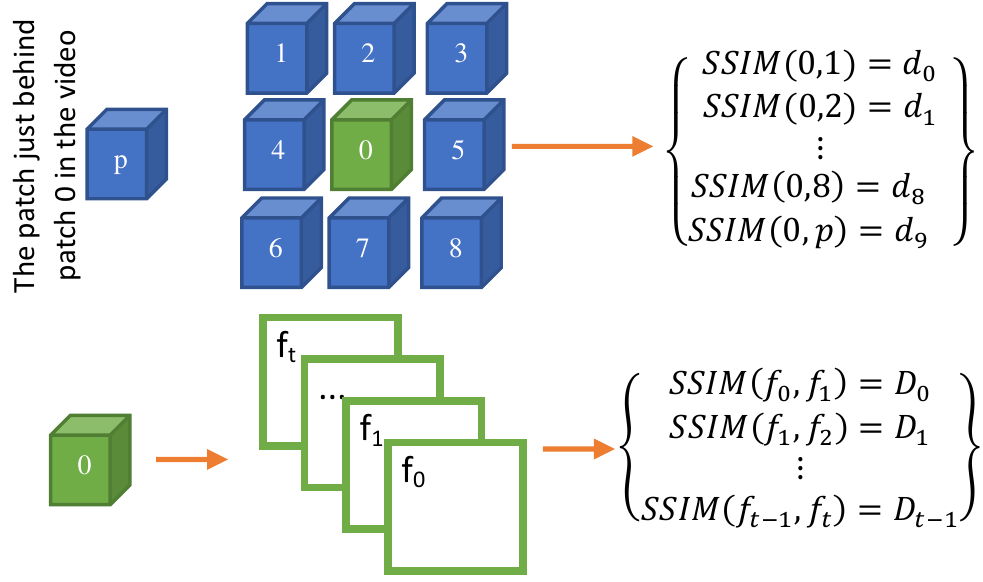}
\end{center}
   \caption{Illustration of our local descriptor: Similarities of each patch on interest with its neighboring patches (top), temporal inner similarities of each patch of interest (bottom).}
\label{fig:3}
\end{figure}

\begin{figure*}[t]
\begin{center}
  \includegraphics[width=0.8\linewidth]{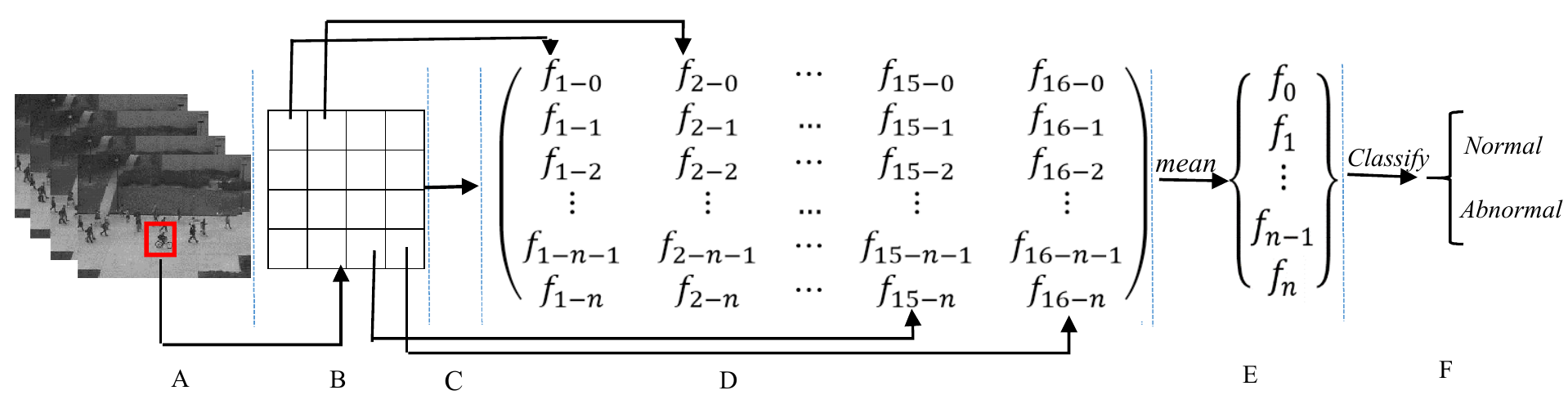}
\end{center}
   \caption{Large patch anomaly detection using feature learning. 
   (A) Input video. 
   (B) Selected test patch (\eg, 40$\times$40$\times$5)  is divided into 16 small patches. 
   (C) $W_{1}\times$ small patch. 
   (D) Pooling all feature vectors (16 vectors). 
   (E) Computing the mean of each feature and create one feature vector.
   (F) Classifying with the learned classifier using 10$\times$10$\times$5 patches.}
\label{fig:4}
\end{figure*}

\vspace{1mm} {\bf Local descriptors.}
To describe each video patch, we use a set of local features. The similarity between each patch and its neighboring patches are calculated. As for the neighbors, we consider nine spatial neighboring patches and one temporal neighboring patch (the one right behind the patch of interest when arranged temporally), yielding to 10 neighbors for each single patch. For temporal neighbors, we only consider the patch before the patch of interest (not the next one), as we aim to detect the anomaly soonest possible, even before the next video frames (and therefore patches) in the video stream arrive. We use SSIM for computing the similarity between two patches, which is a well-known image-quality assessment tool~\cite{BRU2012}. Further, as a second type of local descriptor, we calculate the SSIM of each single frame with its subsequent frame in the patch of interest. Figure~\ref{fig:3} illustrates our local feature assessment through the spatio-temporal neighboring. The local descriptor would be the combination of the SSIM values, \ie, $[d_{0}\cdots d_{9},D_{0} \cdots D_{t-1}]$.

\vspace{1mm} {\bf Anomaly Classifier.}
To model the normal activities in each video patch, we incorporate two Gaussian classifiers $ C_{1}$ and $C_{2}$. For classifying  $x^{\prime}$ patches, as described, we use two partially independent feature sets (global and local), and compute the Mahalanobis distance $f(y)$.
If $f(y)$ is larger than the threshold then it is considered to specify an abnormal patch, where $y$ equals $W_{1}\times x^{\prime}$ in the global classifier, and $[d_{0}\cdots d_{9},D_{0} \cdots D_{3}]$ for the case of the local classifier. To avoid numerical instabilities, density estimates are avoided. As a result, the $C_{1}$ and $C_{2}$   classifiers are defined as follows:
\begin{equation}
C_{i}(x)=
\begin{cases}
{\rm Normal} &  f(x) \le {\rm threshold} \\
{\rm Anomaly} &  {\rm otherwise}
\end{cases}
\end{equation}
with
\begin{equation}
f(x)=(x-\mu)^T\Sigma^{-1}(x-\mu)
\end{equation}
where  $\mu$ and $\Sigma$ are mean and covariance matrix, 
respectively. Selecting a ``good'' threshold is important for the performance; it can be selected 
based on training patches.
As mentioned before, if both  $ C_{1}$ and $C_{2}$ classifiers label a patch as being an anomaly, 
it is considered to be an anomaly, but if one or neither of them considers the patch as being an anomaly, 
our algorithm classifies it as being a normal patch. A summary of these criteria is shown as $F$ function
in the following equation: 
\begin{equation}
F(x)=\begin{cases}
{\rm Normal} &   {\rm if}  \; C_{1} = {\rm Normal} \> \wedge \>C_{2} = {\rm Normal} 
\\
{\rm Anomaly} &  {\rm otherwise} \qquad \qquad \qquad \qquad \;\; (4)
\end{cases}
\nonumber
\end{equation}

\vspace{1mm} {\bf Anomaly detection using feature learning.}
We learn the features from raw training data, 
and classify the video patches as specified in the previous section. 
But based on the idea in \cite{BER2012}, 
using both small patches and large patches usually leads to increased values of false-positive rate 
and decreased value of true-positive rate, respectively. When the patches become larger, 
the input dimension of the auto-encoder increases, so the number of weights in the 
network, which need to be learned, will also increase. 

Under the condition of limited training examples, learning of features from 
large patches is impractical (for example 40$\times$40$\times$5),  to overcome these challenges, 
we learn the features from (small) 10$\times$10$\times$5 patches. To create a model using these 
features, in the test phase the large patches (40$\times$40$\times$5) are considered. 
Because the learned classifier is adapted for 10$\times$10$\times$5 patch representations,   
we convolve the learned feature ($W_{1}$) in 40$\times$40$\times$5 patches, without overlapping, 
and pool the 16 extracted  feature vectors from the 40$\times$40$\times$5 patches. So, we use 
mean pooling to achieve a representation of 40$\times$40$\times$10 patches that can be checked with 
the learned classifier using 10$\times$10$\times$5 patches. This procedure is shown in Figure ~\ref{fig:4}.

\section{Experimental results and comparisons}
\label{sec:ER}

We compare our algorithm with state-of-the-art methods on Ped2 UCSD\footnote{www.svcl.ucsd.edu/projects/anomaly/dataset.htm} and UMN\footnote{mha.cs.umn.edu/Movies/Crowd-Activity-All.avi} benchmarks. We empirically demonstrate that our approach is suitable to be used in surveillance systems.

{\bf Experimental settings.}
Feature learning is done with 10$\times$10$\times$5 patches. Training and testing phases in anomaly detection is done with 10$\times$10$\times$5 and 40$\times$40$\times$5 patch sizes, respectively. In anomaly detection, the size 40$\times$40$\times$5 is exploited. Feature learning is done with an auto-encoder with 0.05 sparsity. Each 10$\times$10$\times$5 patch is represented by a 1000-dimensional feature vector. Before feature learning, normalization is performed to set the mean and variance to 0 and 1, respectively.
   
{\bf UCSD datasets.}
This dataset includes two subsets, ped1 and ped2, that are from two different outdoor scenes. Both are recorded with a static camera at 10 fps, with the resolutions $158\times234$ and $240\times360$, respectively. The dominant mobile objects in these scenes are pedestrians. Therefore, any object (\eg, a car, skateboarder, wheelchair, or bicycle) is considered as being an anomaly. We evaluate our algorithm on ped2. This subset includes 12 video samples, and each sample is divided into training and test frames. To evaluate the localization, we utilize the ground truth of all test frames. We compare our results with state-of-the-art methods using receiver operating curve (ROC) and equal error rate (EER) analysis, similar to~\cite{MAH2010}. We use two evaluation measures, one at frame level and the other at pixel level. In addition to these, we define a new measure for the accuracy of anomaly localization, called {\it dual pixel level}. These measures are defined as follows:

{\it Frame level measure}: If one pixel detects an anomaly then it is considered as being an anomaly.

{\it Pixel level measure}: If at least 40 percent of anomaly ground truth pixels are covered by pixels detected by the algorithm, then the frame is considered to be an anomaly.  
 
Suppose that the algorithm detects some region as being an anomaly, and just one of these regions has an overlap with anomaly ground truth; the number of false regions is not considered in the two former measures. Such a region is called a ``lucky guess''.
For considering the ``lucky guess'' regions, we introduce the dual pixel level. This measure is sensitive to a ``lucky guess''.  

\begin{figure}[b!]
\begin{center}
  \includegraphics[width=0.9\linewidth]{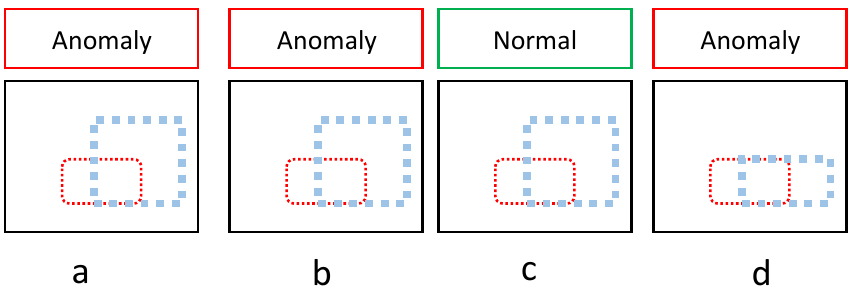}
\end{center}
 \caption{Measure of anomaly evaluation. The blue and red rectangles indicate the 
 output of the algorithm and anomaly ground truth, respectively. 
 (a) Frame-level. 
 (b) Pixel- level evaluation: 40 percent red (ground truth) is covered with blue (detected).
 (c) Dual pixel-level: Evaluates that 40 percent of red is covered by blue, but at least 
 $\beta$ percent of blue is not covered by red.
 (d) Dual-pixel level}
\label{fig:5}
\end{figure}

{\it Dual pixel level}:  In this measure, a frame is considered as being an anomaly if (1) it satisfies the anomaly condition at pixel level and (2) at least $\beta$ percent (\ie, $10\%$) of the pixels detected as anomaly are covered by the anomaly ground truth. If, in addition to the anomaly region, irrelevant regions are also considered as being an anomaly,  then this measure does not identify the frame as being positive.  
Figure~\ref{fig:5} shows an example for the different measures of anomaly detection. 

\begin{figure}[t!]
\begin{center}
  \includegraphics[width=1\linewidth]{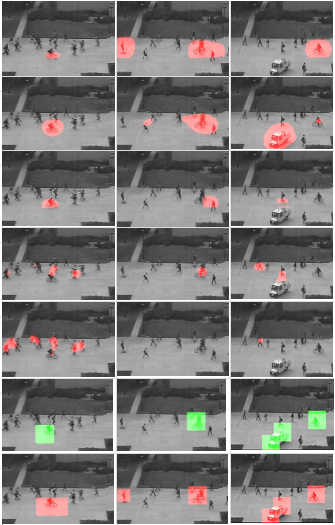}
\end{center}
   \caption{
   Example of anomaly detection from three scenes. 
   First row to 7th row show Temporal MDT, Spatial MDT, MPPCA, Social force, 
   Optic flow, Our method (feature learning only), an Our method (combined views)
   }
\label{fig:6}
\end{figure}

{\bf Performance Evaluations.}
Figure~\ref{fig:6} shows a qualitative comparison with other methods.\footnote
   {
   Our results  are available at http://mahfathy.iust.ac.ir/. 
    }
This figure indicates that our algorithm has the best performance in comparisons with all the competing algorithms. For the run-time comparisons, see Table~\ref{tab:time}.
 
\begin{table}[h!]
\begin{center}
\begin{tabular}{cc}
\hline
Method   &	Time (second per frame)  \\
\hline\hline
Xua \etal \cite{XU2014}& Offline \\
Li \etal \cite{LI2014} & 1.38 \\
Ours & 0.04 \\
\hline
\end{tabular}
\end{center}
\caption{Run time comparison}
\label{tab:time}
\end{table}

\begin{figure*}
\begin{center}
  \includegraphics[width=0.45\linewidth]{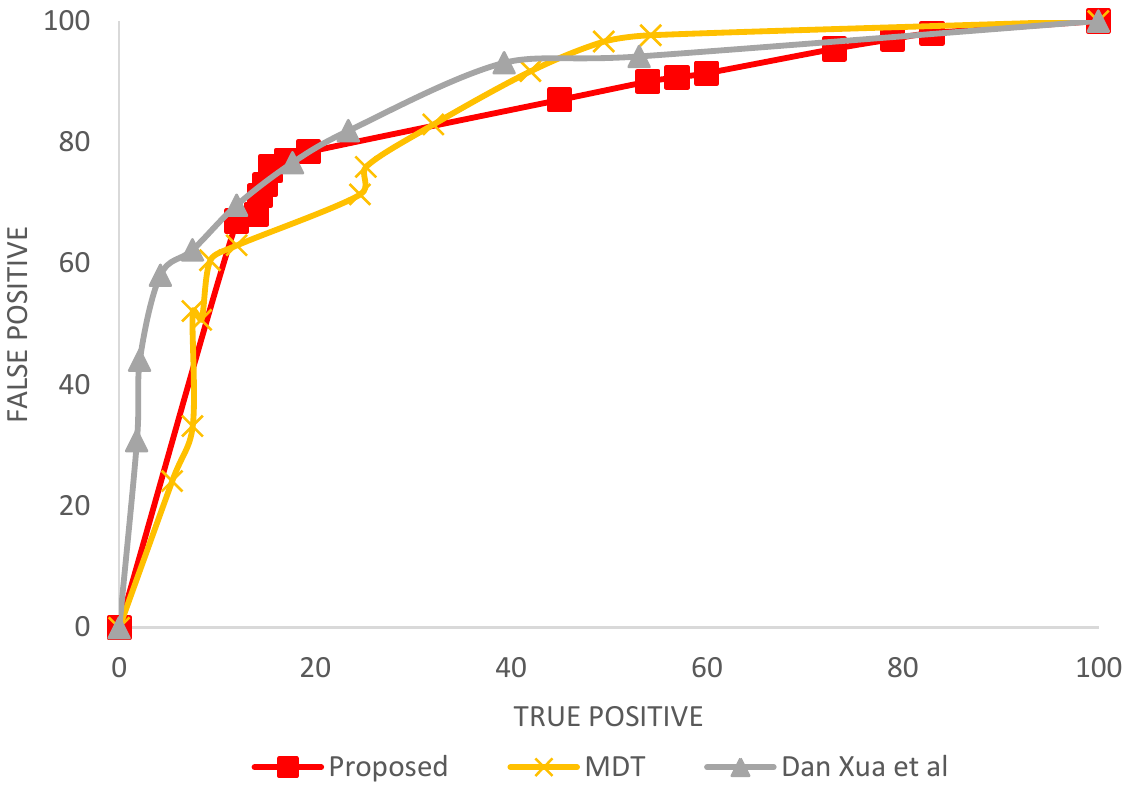}
  \includegraphics[width=0.45\linewidth]{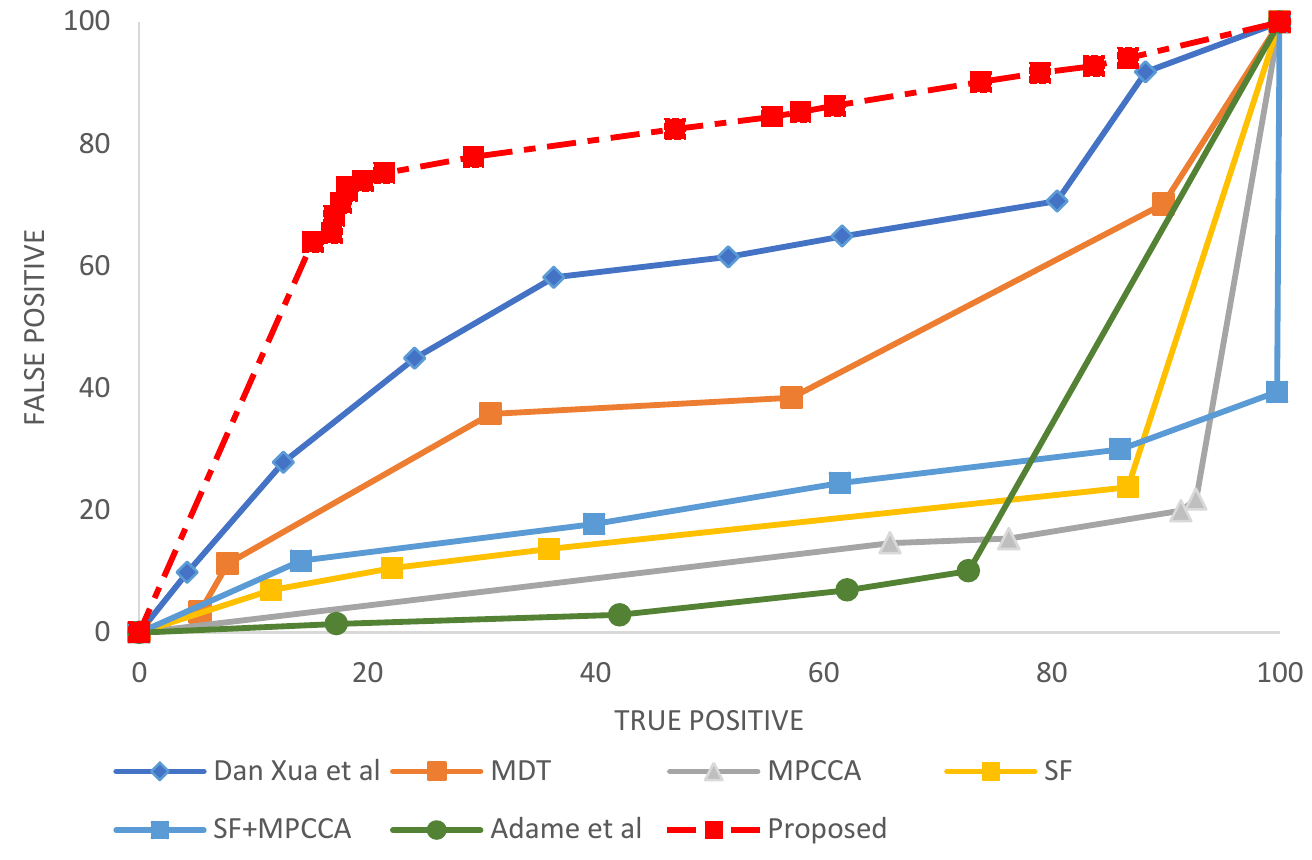}
\end{center}
   \caption{
   Comparison ROC curve ({\it left to right}): 
   Frame-level evaluation and pixel- level evaluation
   }
\label{fig:7}
\end{figure*}

\begin{table}[h!]
\begin{center}
\begin{tabular}{|c|c|c|}
\hline
Method   &	Frame-level  &  Pixel-level   \\
\hline\hline
SF~\cite{MEH2009} & 42 &  79 \\
MPCCA  ~\cite{KIM2009} & 30 &  82 \\
MPCCA+SF ~\cite{MAH2010} & 36 &  72 \\
Adam et.al  ~\cite{ADA2008}& 42 &  76 \\
MDT ~\cite{MAH2010} & 25& 55\\
Xua \etal ~\cite{XU2014} &20&42 \\
Li \etal  ~\cite{LI2014}& 18.5& 29.9 \\
Ours &  19&24 \\
Ours 0.1 &  ---  &  67.5 \\
Ours 0.05 & --- & 27.5 \\
\hline
\end{tabular}
\end{center}
\caption{EER for frame and pixel level comparisons}
\label{tab:EER}
\end{table}

In Figure~\ref{fig:7} (Left), the frame-level ROC of our method is compared with other methods on the ped2 dataset. It shows that our method is comparable to other methods. For this measure, the EER for frame level for different methods is shown in Table~\ref{tab:EER}. This confirms that our method has a good performance in comparison to others. We outperform all of the methods except the one of Li \etal (we are 0.5 percent below), reported in \cite{LI2014}.

\begin{figure}[b!]
\begin{center}
  \includegraphics[width=0.9\linewidth]{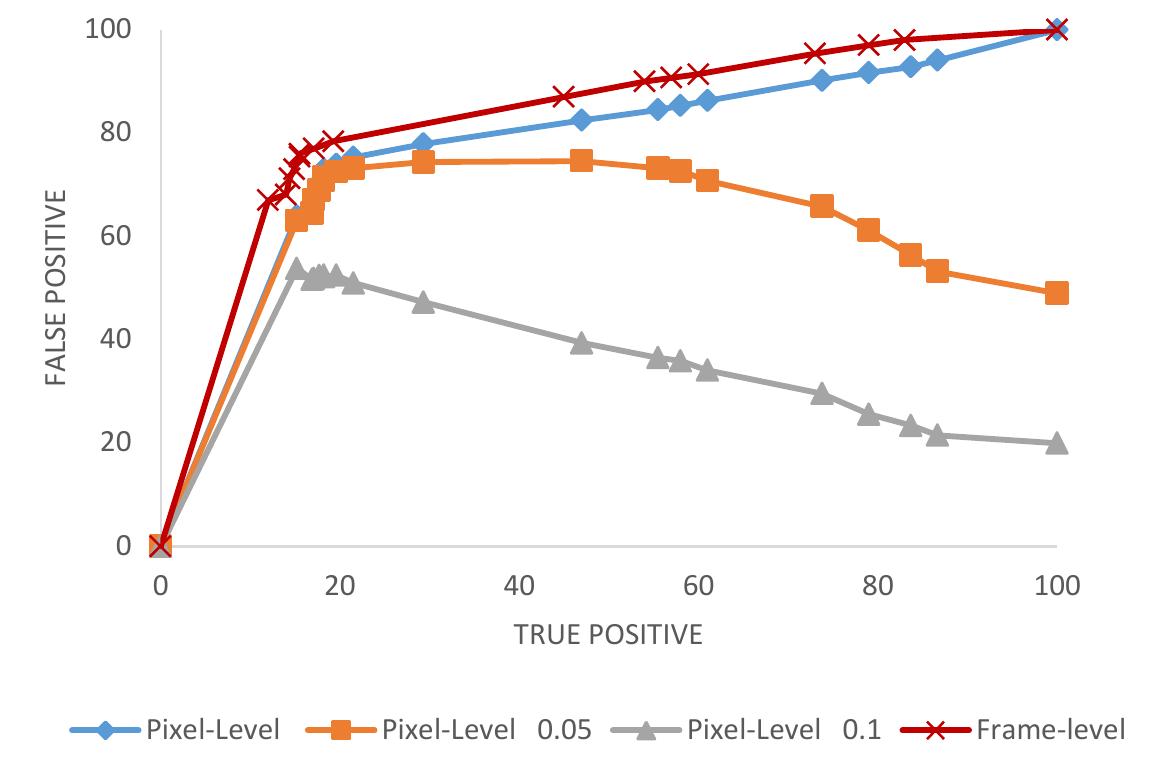}
\end{center}
   \caption
   {
   Comparison between dual pixel localization with $\beta$ equal to 
   0 (pixel- level measure), 0.05, 0.10, and frame-level 
   }
\label{fig:8}
\end{figure}

Figure~\ref{fig:7} (Right) illustrates the ROC with respect to the pixel-level measure. In Table 1, we compare the pixel level EER of our approach to that of other approaches.  Our method's EER is 24  percent where the next best result is 29.9 percent reported for the method Li \etal \cite{LI2014}. Our method is 5.9 percent better than the otherwise best result. 
The results show (both ROC and EER) that our algorithm outperforms the other methods for the pixel-level measure. We also use a dual-pixel level measure to analyze the accuracy of anomaly localization.  Figure~\ref{fig:8} shows the effect of the parameter $\beta$ on our algorithm. The algorithm has a good performance, even better than the state-of-the-art, in pixel level with $\beta$=0.05 percent and 0 percent. Figure~\ref{fig:8} illustrates comparisons at frame level and pixel level of our approach; in contrast to all reported algorithms, the pixel level measure is very close to frame level measure in our algorithm. 

\begin{figure}[h!]
\begin{center}
  \includegraphics[width=1\linewidth]{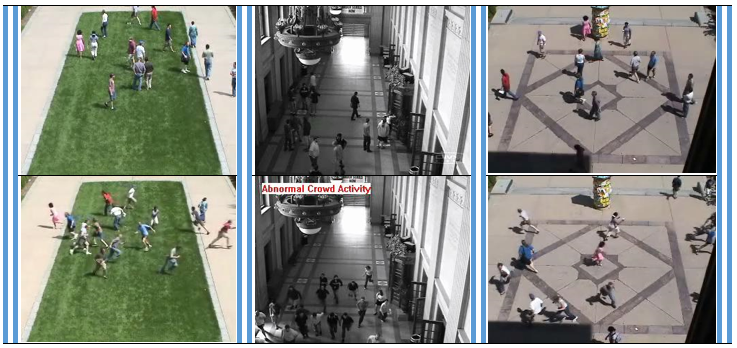}
\end{center}
   \caption
   {Examples of normal and abnormal crowed activities in scenes of the UMN dataset.
   {\it Top}: Normal.  {\it Bottom}:  Abnormal   
   }
\label{fig:UMN}
\end{figure}

\begin{table}[h!]
\begin{center}
\begin{tabular}{|c|c|c|}
\hline
Method   &	EER &  AUC   \\
\hline\hline
Chaotic invariants  ~\cite{WU2010} & 5.3 &  99.4 \\
SF~\cite{MEH2009} & 12.6 &  94.9 \\
Sparse ~\cite{CON2011} &  2.8& 99.6  \\
Saligrama et.al ~\cite{SAL2012} & 3.4&99.5\\
Li \etal  ~\cite{LI2014}& 3.7& 99.5 \\
Ours & 2.5  & 99.6  \\
\hline
\end{tabular}
\end{center}
\caption{Anomaly detection performance in EER and AUC}
\label{tbl:umn}
\vspace{-4mm}
\end{table}

{\bf UMN dataset.}
The UMN dataset has three different scenes. In each scene, a group of people are walking in an area, suddenly all people run away (escape); the escape is considered to be the anomaly. Figure~\ref{fig:UMN} shows examples of normal and abnormal frames of this dataset.

This dataset has some limitations. There are only three anomaly scenes in the dataset, and the temporal-spatial changes between normal and abnormal frames are very high. This dataset has no pixel-level ground truth. Based on this limitations, to evaluate our method, the EER and AUC in frame-level are used. The EER and AUC results are shown in Table~\ref{tbl:umn}. Because this dataset is simple, and anomaly localization is not important, only the global detector is used. Previous methods performed reasonably good on this dataset. The AUC of our method is comparable with the otherwise best result, and the EER of our approach is better (by 0.3 percent) than the one of the best previous method.

\section{Conclusions}
\label{sec:C}

We presented an anomaly detection and localization method. In our method, we propose to represent a video using both global and local descriptors. Two classifiers are proposed based on these two forms of representation. Our fusion strategy on the outputs of these two classifiers achieves accurate and reliable anomaly detection and localization. However, each of the two classifiers has a good performance for anomaly detection, solely. This is especially shown on the UMN dataset where the global descriptor achieves state-of-the-art results. We introduced a new metric for region level anomaly detection for suspicious regions, as well. The performance of our approach on the UCSD dataset is better compared to recent approaches. It is also worth noting that we achieve all these good results in a much better running time than all the competing methods. Our method enjoys a low computational complexity, and can be run in real-time. This makes it quite useful for real-time surveillance applications, in which we are dealing with live streams of videos.

{\footnotesize
\bibliographystyle{ieee}

\begin{thebibliography}{10}
\itemsep=-1pt

\bibitem{ADA2008}
A.~Adam, E.~Rivlin, I.~Shimshoni, and D.~Reinitz.
 Robust real-time unusual event detection using multiple fixed location monitors.
{\em IEEE Trans. Pattern Analysis Machine Intelligence}, 
30(3):555--560, 2008.

\bibitem{BEN2009}
Y. Benezeth, P.-M. Jodoin, V. Saligrama, and C. Rosenberger.
Abnormal events detection based on spatio-temporal co-occurrences.
In {\em CVPR},  
pages 1446--1453, 2009.

\bibitem{BER2012}
M.~Bertini, A.~Del~Bimbo, and L.~Seidenari.
Multi-scale and real-time non-parametric approach for anomaly detection and localization.
{\em Computer Vision Image Understanding}, 
116(3):320--329, 2012.

\bibitem{BRU2012}
D.~Brunet, E.~R. Vrscay, and Z.~Wang.
On the mathematical properties of the structural similarity index.
{\em IEEE Trans. Image Processing}, 
21(4):1488--1499, 2012.

\bibitem{COA2011}
A.~Coates, A.~Y. Ng, and H.~Lee.
An analysis of single-layer networks in unsupervised feature learning.
In {\em Int. Conf. Artificial Intelligence Statistics}, 
pages 215--223, 2011.

\bibitem{CON2011}
Y.~Cong, J.~Yuan, and J.~Liu.
Sparse reconstruction cost for abnormal event detection.
In {\em CVPR},
pages 3449--3456, 2011.

\bibitem{CON2013}
Y.~Cong, J.~Yuan, and Y.~Tang.
Video anomaly search in crowded scenes via spatio-temporal motion context.
{\em IEEE Trans. Information Forensics Security},
8(10):1590--1599, 2013.

\bibitem{JIA2011}
F.~Jianga, J.~Yuan, S. A.~Tsaftarisa, and A. K.~Katsaggelosa.
Anomalous video event detection using spatiotemporal context.
{\em Computer Vision Image Understanding}, 
115(3):323--333, 2011.

\bibitem{KIM2009}
J.~Kim and K.~Grauman.
Observe locally, infer globally: a space-time MRF for detecting abnormal activities with incremental updates.
In {\em CVPR}, 
pages 2921--2928, 2009.

\bibitem{KRA2009}
L. Kratz and K. Nishino.
Crowded scenes using spatio-temporal motion pattern models.
In {\em CVPR},  
pages 1446--1453, 2009.

\bibitem{LI2014}
W.~Li, V.~Mahadevan, and N.~Vasconcelos.
Anomaly detection and localization in crowded scenes.
{\em IEEE Trans. Pattern Analysis Machine Intelligence}, 
36(1):18--32, 2014.

\bibitem{LU2013}
C.~Lu, J.~Shi, and J.~Jia.
Abnormal event detection at 150 fps in MATLAB.
In {\em ICCV}, 
pages 2720--2727, 2013.

\bibitem{MAH2010}
V.~Mahadevan, W.~Li, V.~Bhalodia, and N.~Vasconcelos.
Anomaly detection in crowded scenes.
In {\em CVPR}, pages 1975--1981, 2010.

\bibitem{MEH2009}
R.~Mehran, A.~Oyama, and M.~Shah.
Abnormal crowd behavior detection using social force model.
In {\em CVPR}, 
pages 935--942, 2009.

\bibitem{ROS2013}
M.~J. Roshtkhari and M.~D. Levine.
An on-line, real-time learning method for detecting anomalies in
  videos using spatio-temporal compositions.
{\em Computer Vision Image Understanding}, 
117(10):1436--1452, 2013.
  
\bibitem{SAL2012}
V.~Saligrama and Z.~Chen.
Video anomaly detection based on local statistical aggregates.
In {\em CVPR}, 
2012.

\bibitem{VIN2008}
P.~Vincent, H.~Larochelle, Y.~Bengio, and P.-A. Manzagol.
Extracting and composing robust features with denoising autoencoders.
In {\em Int. ACM Conf. Machine
  Learning}, pages 1096--1103, 2008.
  
\bibitem{WAN2007}
X. Wang, X. Ma, and E. Grimson.
Perception by hierarchical Bayesian models. 
In {\em CVPR}, 
pages 1--8, 2007.
  
\bibitem{WU2010}
S.~Wu, B.~Moore, and M.~Shah.
Chaotic invariants of Lagrangian particle trajectories for anomaly detection in crowded scenes.
In {\em CVPR}, 
2010.

\bibitem{XU2014}
D.~Xu, R.~Song, X.~Wu, N.~Li, W.~Feng, and H.~Qian.
Video anomaly detection based on a hierarchical activity discovery within spatio-temporal contexts.
{\em Neurocomputing}, 
143:144--152, 2014.

\bibitem{YAN2013}
Y.~Yang, G.~Shu, and M.~Shah.
Semi-supervised learning of feature hierarchies for object detection in a video.
In {\em CVPR}, 
pages 1650--1657, 2013.

\bibitem{ZHA2005}
D.~Zhang, D.~Gatica-Perez, S.~Bengio and I. McCowan.
Semi-supervised adapted HMMS for unusual event detection.
In {\em CVPR}, 

\hspace{6mm} 2005.

\end{thebibliography}

}

\end{document}